\documentclass[PHME, 2022]{PHMSociety}

\usepackage{graphicx}
\usepackage{amsmath}
\usepackage{adjustbox}
\usepackage{tikz}
\graphicspath{{Figures/}} 
\usepackage{pgfplots}
\DeclareUnicodeCharacter{2212}{−}
\usepgfplotslibrary{groupplots,dateplot}
\usetikzlibrary{patterns,shapes.arrows}
\pgfplotsset{compat=newest}

\usepackage{xurl}

\RequirePackage{color}

\begin{document}

\title{Machine Learning Methods for Health-Index Prediction in Coating Chambers}

\author{%
	Clemens Heistracher\authorNumber{1}, Anahid Jalali\authorNumber{2}, Jürgen Schneeweiss\authorNumber{3}, Klaudia Kovacs\authorNumber{4},  Catherine Laflamme\authorNumber{5} and Bernhard Haslhofer\authorNumber{6}
}

\address{

	\affiliation{{1,2}}{AIT Austrian Institute of Technology, Vienna, 1210, Austria}{ 
		{\email{Clemens.Heistracher@ait.ac.at}}\\ 
        {\email{Anahid.Jalali@ait.ac.at}}
		} 

	\tabularnewline 

	\affiliation{3}{D. Swarovski KG, Wattens, 6112, Austria}{ 
		{\email{Juergen.Schneeweiss@swarovski.com}}
		} 

	\tabularnewline 

	\affiliation{4,5}{Fraunhofer Austria Research , Vienna, 1040, Austria}{ 
    {\email{catherine.laflamme@fraunhofer.at}}\\ {\email{klaudia.kovacs@fraunhofer.at}}
	} 

	\tabularnewline 

	\affiliation{6}{Complexity Science Hub, Vienna, 1080, Austria}{ 
		{\email{haslhofer@csh.ac.at}}
		} 
}

\maketitle
\pagestyle{fancy}
\thispagestyle{plain}

\phmLicenseFootnote{Clemens Heistracher}

\begin{abstract}

Coating chambers create thin layers that improve the mechanical and optical surface properties in jewelry production using physical vapor deposition.
In such a process, evaporated material condensates on the walls of such chambers and, over time, causes mechanical defects and unstable processes. As a result, manufacturers perform extensive maintenance procedures to reduce production loss. 
Current rule-based maintenance strategies neglect the impact of specific recipes and the actual condition of the vacuum chamber. Our overall goal is to predict the future condition of the coating chamber to allow cost and quality optimized maintenance of the equipment. This paper describes the derivation of a novel health indicator that serves as a step toward condition-based maintenance for coating chambers. We indirectly use gas emissions of the chamber's contamination to evaluate the machine's condition. Our approach relies on process data and does not require additional hardware installation. 
Further, we evaluated multiple machine learning algorithms for a condition-based forecast of the health indicator that also reflects production planning. Our results show that models based on decision trees are the most effective and outperform all three benchmarks, improving at least $0.22$ in the mean average error.  
Our work paves the way for cost and quality optimized maintenance of coating applications.

\end{abstract}

\section{Introduction}
\label{intro}

Thin-film coatings can manipulate the optical properties of smooth surfaces and create color effects and unique reflection properties in jewelry. Metallic or dielectric layers create color effects and define the reflection coefficient. Such layers are created through the deposition of vaporized material in a vacuum chamber. These processes require high stability since these effects require layer thickness in the order of the wavelength of the optical spectrum. A typical layer thickness ranges from 100 nm to several micrometers.
Machine operators perform regular maintenance and cleaning of the production equipment to guarantee process stability and product quality. The removal of deposits on the vacuum chamber's wall is the primary goal of these activities. 

Maintenance operations are commonly scheduled based on the number of runs since the last procedure. However, the deposits on the walls and their impact on the process depend strongly on the materials and recipes used. Thus, process engineers expect significant cost savings from a maintenance schedule based on the actual condition of the vacuum chamber.
An ideal maintenance schedule would perform all required operations to prevent failures while minimizing operations to save costs.
However, such a predictive maintenance approach requires knowledge of the current and especially the future condition of the vacuum chamber, which is not known for our application.
Therefore, we propose a novel method to estimate the health condition of the vacuum chamber.

Several studies have already focused on 
Health-Index assessment for industrial assets in manufacturing \cite{khoddam2016performance}, semiconductor productions \cite{djeziri2015health} 
and specially for vacuum equipment such as vacuum pumps \cite{7930066}.

However, none of those approaches combines the Health-Index assessment with a forecasting model and uses it for maintenance optimization in coating chambers. We aim to predict the future health status of a coating chamber to optimize the scheduling of maintenance activities. 
To reach this goal, we performed an extensive exploratory data analysis and modeling and can summarize our contributions as follows:
\begin{enumerate}
    \item We derived a Health-Index to describe the degradation in a coating chamber
    \item We evaluate various shallow- and deep learning models for health index prediction on real production data 
\end{enumerate}

Our work provides insights into creating a Health-Index for vacuum chambers that require periodic maintenance and serves as a guideline in model selection for machine learning practitioners and process engineers.
In the following, we briefly introduce related background on coatings with thin layers and the forecasting of a Health-Index (Section \ref{Background}). Then in Section \ref{Method}, we present our data set, exploratory data analysis, and the method of creating a Health-Index, before describing our experiments in Section \ref{results} and our results in Section \ref{results}.
\section{Background}
\label{Background}
\subsection{Decorative Coatings}

Physical vapor deposition (PVD) is the method of choice to create color effects for decorative elements \cite{REINERS199433}. It produces thin layers on surfaces that form wave interferences with the reflected light, influencing the perceived color and reflection. These effects can be seen in soap bubbles or thin oil films on water. The demand for the layer's thickness precision is high as it has to be far below the typical optical wavelength to create a consistent color effect and requires extensive process control.
The industrial production, multiple articles are treated in parallel in coating chambers. The primary working mechanism of such a chamber is the evaporation of substrate, which then condenses on the target. Due to their refraction coefficient, aluminum, zinc, and titanium are suitable choices \cite{jehn1992decorative}. Evaporation occurs in a vacuum to avoid collisions, oxidation, and surface contamination. Additional ion bombardment improves the mechanical properties of the surface. \cite{coatings8110402}

\subsection{Data-Driven Predictive Maintenance Approaches}
Through the advancement of technology and specifically since the introduction of industry 4.0 with the core concept of building smart factories, production, and logistics, the strategies for Predictive Maintenance gained more popularity~\cite{zhang2019data}.  
PdM approaches can be grouped into three; model-based, knowledge-based, and data-driven \cite{ZONTA2020106889}. We narrow the scope and only focus on state of the art for data-driven approaches. A group of these studies focuses on data pre-processing and more feature engineering approaches, combined with shallow and statistical modeling. For example, \cite{umeda2021planned} proposed a maintenance schedule updater based on probabilistic variability of the Remaining Useful Life (RUL) and maintenance costs, which is component agnostic.
\cite{chien2020data} used Partial Least Square supervised learning to model the fault detection and classification parameters of glass substrates for Thin Film Transistor Liquid-Crystal Display (TFT-LCD) manufacturing process. 
Other approaches use complex modeling techniques, such as Deep Neural Networks (DNNs), to predict and analyze equipment failures. \cite{sateesh2016deep} were the first to use Convolutional Neural Networks for the RUL estimation in turbine engines. \cite{7927458} showed that restricted Boltzmann machines could effectively predict RUL in gears.  \cite{8815721} studied the RUL estimation for lithium-ion batteries using long short-term memory-based neural networks. \cite{8439898} build a  neural network model for predictive maintenance of photovoltaic panels. 

Predicting a health indicator (HI) is a related research area. 
Statistical analysis was performed to derive a HI for power transformers \cite{MURUGAN2019274}, transmission lines \cite{6559646}, bearings \cite{PAN2009669}, electrical machines \cite{7377088}, semiconductor production \cite{5170100} and bridges \cite{dohler2014structural}.







\section{Method}
\label{Method}

Our work aims to provide a guideline for implementing a predictive maintenance system for coating chambers in jewelry production. This section presents the details of the real-world dataset and outlines all significant steps in creating a health index and a time-series prediction model.

\subsection{Dataset}
\begin{figure}[t]
\centering
\input{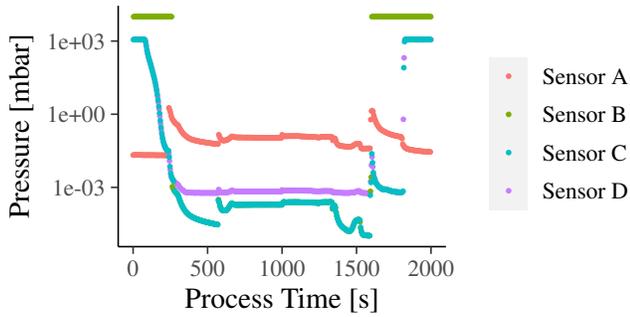}

    \caption{Pressure curves of a single run recorded on four sensors on a logarithmic scale.}
    \label{fig:pressire_cure}
    
\end{figure}

We use data from the real-world production of decorative elements obtained in 15 months. The dataset consists of sensor recordings and process parameters of five assets for the deposition of optical layers. The sensor data consists of recordings directly from the process chamber, which are used to control the conditions in the chamber during production. It contains temperature,  pressure, gas flow recordings, and the electrical parameters of evaporators and ion sputter components. Typically, the process parameters are the settings of equipment that describe the desired conditions, such as the active duration of an ion source or the target pressure. We will refer to these sets of rules as recipes since they are distinct for a type of product and color effect. Additionally, the dataset contains maintenance-related information, such as the number of runs since the last cleaning.
The vacuum system consists of three parts: The initial stage is a backing pump with an operating range at atmospheric pressure. Then, a turbomolecular pump starts at vacuum pressures, and a cold trap is activated. These stages are monitored and controlled by four pressure sensors in the vacuum chamber due to the high range of pressures from $10^3 mbar$ to  $10^{-4} mbar$ and the limited measuring range for single sensors. 
Figure \ref{fig:pressire_cure} shows the available pressure curves for a single production run.

Our dataset consists of 2137 production runs of five assets containing sensor measurements for every 0.5 seconds. In addition, we recorded 119 numerical columns and 13 categorical columns containing product and recipe information and timestamps. Thus, the whole dataset contains approximately 10 million samples of 134 features.

\subsection{Health-Index Derivation}
\label{sec_method_hi}
Health indicators describe equipment conditions from a maintenance perspective and are used to implement adequate repair and service activities. A typical health index for coating chambers is the deposit accumulation in the vacuum chamber \cite{8666799}, which can be measured by piezo-based sensors \cite{BENES19951}. However, this approach only measures the layer thickness and does not consider the potential interaction of different layer materials, recipes, and unknown effects. Domain experts assume that layers of alternating materials impact the stability of the deposit and its ability to outgas and thus the condition of the chamber. We believe that the effect of alternating materials can be learned by models when provided with recipe information and the corresponding sensor data. Thus, our goal is to consider different materials' impact and potential interaction, which requires further investigation, and therefore, we require a HI derived from the current measurements.

We based the development of our health indicator on domain experts' observation that the pumping duration correlates with the condition of the chamber. Additionally, \cite{field2016reduced} supports this assumption and shows a correlation between pumping duration and the quality of produced coatings for laser applications. 
Before and after every production run, operators open the vacuum chamber to load or unload the products, and air at atmospheric pressure fills the chamber while the door is open. Multi-stage vacuum pumps evacuate the chamber before every coating procedure, and domain experts have noticed that this pumping takes longer the more contaminated the chamber is. 
 
We assume that the pumping duration corresponds to the condition of the vacuum chamber and apply a variety of models and visual analyses to validate this assumption. First, we selected an asset and time frame that uses a single standard recipe to rule out recipe-dependent factors. Our subset for this analysis consists of the pressure curves from four sensors for 400  runs and their corresponding maintenance information. Then, we identified several pressure intervals ($\Delta p1$....$\Delta pn$) based on actual steps in the production process. For instance, the first interval starts at atmospheric pressure, which is the condition at the beginning of the process, and ends at $0.02$ mbar, the pressure at which the turbopump is activated. 
We further extract the processing time of all pressure intervals and create variables $Ti$ with $i \  \epsilon \ [1,2,3,4]$ that correspond to the time it took the pumps to evaluate each interval. For instance, $T1$ is the process duration from atmospheric pressure to $0.02$ mbar. Then, we merged the data with the maintenance information $n_{runs}$, which is the number of runs since the last maintenance procedure. 
Further, we built models to understand the correlation between maintenance conditions and pumping time. In detail, we fit a linear regression model to the data of each pressure range by using the number of runs as input and the pumping time as the target variable. The regression is of the form: 

\begin{equation}
    Ti = k_{i} * n_{runs} + d_i
\end{equation}

$k_i$ is the slope of the regression in $[seconds/run]$  for the ith segment, and $d_i$ is an additive constant. We defined an impact variable that indicates the relative change of the pumping duration over a complete cleaning cycle consisting of 100 runs.
The impact variable is defined as $\alpha_{i} = k_{i}/\overline{t}_{pump_i}*100$, with $\overline{t}_{pump_i}$ being the average pumping duration for a clean machine, which is defined as the mean pumping duration for the first 10 runs after cleaning for the ith segment over all cycles. We quantified the fit based on the coefficient of determination, which roughly gives the proportion of variance that can be explained by the model and the impact variable.

\subsection{Health-Index Forecasting}

\begin{figure}
    \centering
    \input{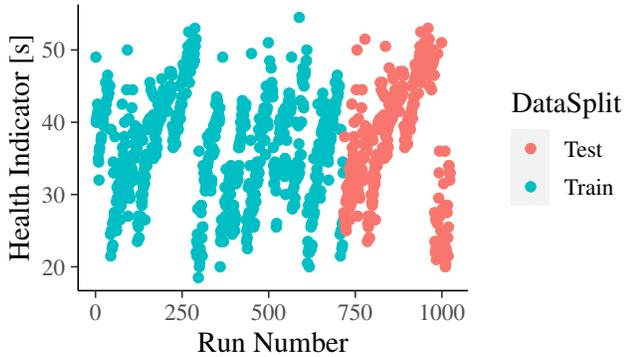}
    \caption{The target variable for a period of 12 months.The colour indicates the train set (blue) and test set (red)}
    \label{fig:T2_curvve}
\end{figure}

Maintenance planners require estimates of the future evolution of an asset to provide adequate resources when needed. This task aims to forecast the future contamination of coating chambers.
Our approach utilizes the health index (HI) we derived in section \ref{sec_method_hi}, which measures the contamination of a vacuum chamber based on the time it takes to evacuate. We predict the future HI based on the process data and knowledge of the planned recipes available at the current run, which can be seen as a regression problem. We predict the value of the HI ten runs after the current run, which roughly corresponds to a day of production and is the typical time available for production and maintenance planning in our application. We use the recipes in the subsequent ten runs in ten one-hot encoded features. Each feature characterizes the recipe used in a future run.   The HI over the period of a year is illustrated in Figure \ref{fig:T2_curvve}.


We aggregate the process data by calculating each run's mean, minimum, maximum, and standard deviation and use one-hot encoding to encode categorical features such as recipes. We split the dataset into a train and test set to evaluate our model on an independent dataset.
We train with the oldest 70\% of the data and test with the most recent $30\%$. 
We selected a range of machine learning models that were used in related tasks such as Decision Trees (DT), Random Forest (RF), K-Nearest Neighbors (KNN), Multilayer Perceptrons (MLP), and Recurrent Neural Networks with Long Short-Term Memory units (LSTMs). 

We evaluate our models using the mean average error (MAE) calculated on the prediction and the actual value of the HI.  
We design benchmarks based on naive assumptions to put our predictions into perspective. Benchmark 1 (BM1) uses the current values as the prediction. 
We calculate the average curve of a cleaning cycle in the train set and use it as benchmark 2 (BM2). Benchmark 3 (BM3) uses the average of all data points in the train set.
\section{Results}
\label{results}

In the following, we present our results for the derivation of a HI for coating chambers and a forecasting model to predict the future condition of the chamber.

\subsection{Health-Index Derivation}

The main goal of this task is to derive a HI  from the process data, which describes the condition of the coating chamber. We observe that the overall process duration increases with the contamination of the vacuum chamber. We use the pumping duration for various pressure steps and model the maintenance data to determine the impact on the production of each step. Table \ref{tab_regression2} shows the coefficient of determination ($R2$) of our models, and the impact variable $\alpha_{i}$. R2 indicates how much of the variation can be explained by our model. The pumping duration for segment $\Delta p2$ has the highest correlation with contamination at an $R2$ score of $0.61$ and an increase of $55\%$ over one cleaning cycle. 
Therefore, we propose the pumping duration of $\Delta p2$ as a new health indicator for coating chambers and will refer to it as T2.

\begin{table}[t] \small  
	\begin{center}  
\caption{Regression of pumping duration and chamber contamination}
\begin{tabular}{ l | l | l  | l | l | l}
		\hline \hline
		\textbf{$i$} & \textbf{$Segment$}	& \textbf{$k_{i} [s / run]$}   	& \textbf{$\overline{t}_{pump} \ [s]$}	& \textbf{$\alpha_{i} [1/cycle]$}	&\textbf{$R2$}            \\ 
		\hline \hline
1&$\Delta p1$ & 0.06              & 139                                       & 5\%                                      & 0.19 \\\hline
2&$\Delta p2$& 0.12             & 21                                        & 55\%                                     & 0.61 \\\hline
3&$\Delta p3$& 0.45             & 285                                       & 28\%                                     & 0.10 \\\hline
4&$\Delta p4$ & 0.85             & 305                                       & 17 \%                                    & 0.11 \\\hline
5&$\Delta p5$ & 0.19              & 160                                        & 12 \%                                     & 0.39 
\end{tabular}
\label{tab_regression2}

\end{center}
\end{table}

\begin{figure}

    \centering
    \input{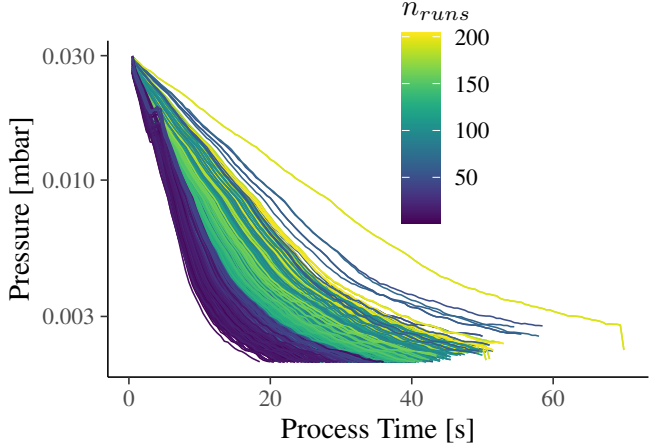}

               \caption{The aligned pressure curves for the segment $\Delta p2$. $n_{runs}$ is indicated by the colour.  }

        \label{fig:T2s}
\end{figure}

In Figure \ref{fig:T2s}, we illustrate a high number of pressure curves for the segment  $\Delta p2$. We align the pressure curves when they reach the upper limit of $\Delta p2$ at $0.03 \ mbar$, and clip the curves at the lower limit at $0.002 \ mbar$. 
Different colors indicate the maintenance conditions, which are the number of runs since the last cleaning cycle.
We observe that the pressure curves decrease with the contamination of the chamber. 

Our results show that the contamination of the chamber can be determined by the pressure curves of the production process, and therefore, the pumping duration is a suitable HI for a coating chamber.

\subsection{Health-Index Prediction}

The main goal of this task is to predict the future health indicator based on information available at the current run. We evaluate the well-established machine learning models such as Support Vector Machines (SVM) \cite{wan2014predictive}, Decision Trees (DT), and Random Forests (RF) \cite{scheibelhofer2012methodology}, and Neural Network architectures optimized for time series \cite{bruneo2019use}. We evaluate our models on an independent dataset, which consists of the most recent $30\%$ of data, and compared them with the previously defined benchmarks. 

Our results, summarized in Table \ref{table_r_preder}, indicate that shallow machine learning models significantly outperform neural network-based approaches and, therefore, are better suited for this task.
The DT achieved the best mean average error (MAE) with $3.0$, followed by SVM with a score of $3.26$ and RF with $3.28$, outperforming all three benchmarks.
However, the naive prediction of BM1 scored $3.22$ and is only slightly outperformed by the best model and outperformed all others.

\begin{table}[t] \small  
	\begin{center}  
		
	\caption{Results forecast health indicator.				\label{table_r_preder}
}

	\begin{tabular}{ l | l  | l | l | l}
		\hline \hline
		\textbf{model}	& \textbf{MAE}  	& \textbf{BM2}	& \textbf{BM1}	& \textbf{BM3}            \\ 
		\hline \hline
SVC &   3.26 &  9.0 &  3.22 &  8.37 \\ \hline
   DT &   3.00 &  9.0 &  3.22 &  8.37 \\\hline
  RF &   3.28 &  9.0 &  3.22 &  8.37 \\\hline
   KNN &   3.81 &  9.0 &  3.22 &  8.37 \\\hline
 MLP &   5.56 &  9.0 &  3.22 &  8.37 \\\hline
 LSTM &  10.03 &  9.0 &  3.22 &  8.37 \\\hline
	\end{tabular}
	\end{center}

\end{table}

\begin{figure}
    \centering
   \input{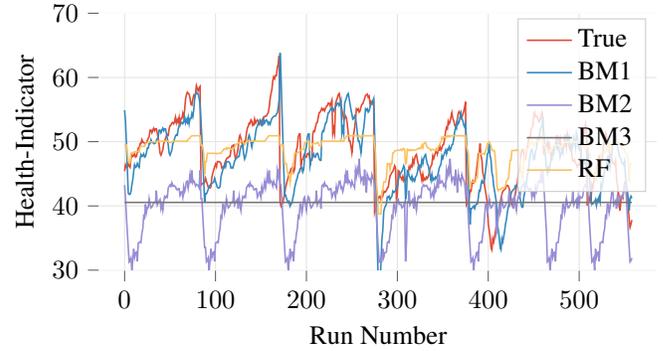}
    \caption{The target variable, prediction and benchmarks of our HI.}
    \label{fig:T2s_bm}
\end{figure}

We illustrate the results of our best models for several cleaning cycles in Figure \ref{fig:T2s_bm}. The shape of the target variable resembles a sawtooth function with a linear ascend during production and a sudden drop after maintenance, which the model appears to have learned.
Benchmarks 2 and 3, learned from the train set, are shifted towards lower values, which explains the poor performance of these benchmarks. Seasonal patterns in temperature and humidity are explanations provided by domain experts. Since the available data spans one year only, we cannot confirm or reject this thesis.


\section{Discussion}
\label{discussion}

A goal of cost-effective maintenance planning is to optimize the timing and type of scheduled activities. This requires knowledge of the current and future condition of the asset, which is often not available. Our work aims to provide this information through data analytics and machine learning models using process data.
This paper presents the derivation of a health indicator that corresponds to the contamination of a coating chamber and a forecasting model to predict the future health indicator based on current process data and planned recipes.  
We based the health indicator development on the domain experts' observation that the time to evacuate the vacuum chamber correlates with its contamination. The adhesion of water on the contaminated walls that evaporates at low pressures is a possible explanation for this behavior. We analyzed the impact of the contamination on the pumping time for various pressure ranges and identified a pressure region that strongly correlates with the cleaning cycle. Our results indicate that this pumping duration could serve as a health indicator for coating chambers. 
Further, we developed predictive models that take the current condition of the chamber and the planned recipes into account and give a prognosis on the future development of the health indicator.

One limitation of our work is the connection between the chamber's performance and our health indicator. We showed that our health indicator correlates to the contamination of the chamber, but we did not provide evidence that it impacts the product quality or process stability. Although we believe it is plausible that outgassing in the vacuum chamber has a negative impact on production, this is an assumption that needs to be verified. \cite{ito2008reduction} showed that dehydration of deposits reduced the number of defects for plasma etching equipment, and we believe that this is transferrable to coating chambers. 
We compared multiple machine learning models to predict our health indicator's future development and found that neural network-based architectures performed poorly compared to traditional machine learning approaches. However, the size of our data set could limit the full potential of neural networks since it is small compared to typical applications of neural networks, and we believe that a larger data set could improve their performance. 

Subsequentially, we can identify two challenges to a profound HI forecasting model. First, the actual impact of our health indicator on product quality must be evaluated. Our dataset originates from a preventive maintenance regime that averts most defects and therefore does not allow us to confirm the benefit of a prolonged maintenance cycle using our approach. Therefore, we need to lift the restrictions on the number of runs and evaluate our approach during actual production with the potential risk of creating defective parts. 
Second, the data set must be extended to include more assets of various designs and more data points. Finally, only results for multiple chamber types on a high number of samples will allow for results that show that our approach is generally valid.


\section{Conclusion}
\label{conclusion}

The lack of methods to assess the actual condition of an industrial asset hinders the industry from moving away from regular maintenance to a more proactive approach. This paper followed the idea that already available process data contains enough information to derive a health indicator, and no additional hardware is required. We address the need for predictive maintenance approaches that are easy to deploy and can be scaled up quickly to many assets and gained some fundamental insight into the current possibilities and most promising future directions. 
In the coming months, we will evaluate the impact of our approach on the actual product quality during production.

\section*{Acknowledgment}
This work was partly funded by the Austrian Research Promotion Agency (FFG) through the project COGNITUS (Project
ID: 3323904).



\bibliographystyle{apacite}
\PHMbibliography{ijphm}

\end{document}